\documentclass[runningheads]{llncs}
\usepackage[T1]{fontenc}
\usepackage{graphicx}
\usepackage{booktabs}
\usepackage[misc]{ifsym}
\newcommand{\corr}{(\Letter)}
\usepackage{mwe}
\usepackage{amsmath}
\usepackage{amssymb}
\usepackage{subcaption}
\usepackage{hyperref}
\usepackage{xcolor}
\usepackage{siunitx}
\usepackage{tcolorbox}
\usepackage{pifont}
\usepackage{multirow}

\begin{document}

\title{Unequal Verdicts: Investigating Gender Bias in LLM-Based Fake News Detection}

\titlerunning{Investigating Gender Bias in LLM-Based Fake News Detection}

\author{Razieh Chalehchaleh\inst{1,2} \corr \and
Reza Farahbakhsh\inst{1} \and
Noel Crespi\inst{1}}
% Use \corr to indicate the corresponding author. Note the spacing around the \corr command. Only one author can be the corresponding author.
% You may leave out the orcidID information, if you want to.

\authorrunning{R. Chalehchaleh et al.}
% If there are two authors, write 'A.L. Benjamin and C.C. Broadus Jr.'
% If there are more than two authors, '[...] et al.' is used.

\institute{
SAMOVAR, Télécom SudParis, Institut Polytechnique de Paris,
91120 Palaiseau, France\\
\email{razieh.chalehchaleh@telecom-sudparis.eu}\\
\email{reza.farahbakhsh@it-sudparis.eu}\\
\email{noel.crespi@it-sudparis.eu}
\and
Arlequin AI, Paris, France\\
\email{razieh@arlq.ai}
}
\maketitle              % typeset the header of the contribution

\begin{abstract}
Large Language Models (LLMs) are increasingly used for automated fact-checking, yet their susceptibility to gender bias in this context remains underexplored.
This study presents the first systematic investigation of gender bias in LLM-based fake news detection using real-world data.
We augment the LIAR benchmark with three gender variants of speaker job titles (\emph{Neutral}, \emph{Male}, \emph{Female}) for each statement to test whether veracity judgments vary solely based on gender presentation.
Six state-of-the-art LLMs are evaluated across multiple bias and fairness metrics.
All models exhibit gender sensitivity: 9.79\%--35.13\% of statements receive inconsistent labels across the three variants, with \emph{Male}--\emph{Female} comparisons showing 6.5\%--23.6\% flip rates.
Two primary bias manifestations are identified: instability (inconsistent judgments) and directionality (systematic favoritism).
Five models show statistically significant directional effects, with the strongest effects displaying male-skeptic patterns.
These findings demonstrate that gender bias undermines both reliability and fairness in LLM-based fake news detection, highlighting the need for bias-aware evaluation and mitigation strategies.
The augmented dataset is publicly released to support future research.

\end{abstract}

\section{Introduction}
The rapid spread of misinformation on social media, particularly during high-impact events, has raised serious concerns due to its potential to harm individuals and societies \cite{rocha2021impact,sanchez2024disinformation,altoe2024online,ecker2024misinformation}.
Extensive work has explored fake news detection and mitigation strategies \cite{alghamdi2024comprehensive,hu2025overview}.
Large Language Models (LLMs) are increasingly applied to fake news and misinformation detection, particularly for automatic verification of claims and news articles \cite{chen2024combating,kumar2024silver,kuntur2024under,yi2025challenges,huang2025unmasking}. 
These models hold the potential to transform how information is verified; however, to harness this potential responsibly, it is essential to critically examine their limitations and remain aware of the risks they may introduce.
Public trust in information is highly influenced by the availability of verification and fact-checking mechanisms on social media platforms \cite{olan2024fake}.
However, inaccurate labeling can backfire: LLM-generated fact checks that misclassify true statements as false have been shown to reduce belief in true news and, conversely, increase belief in deceptive headlines \cite{deverna2024fact}.

LLMs are pre-trained on vast corpora of real-world text, which enables their strong performance \cite{yang2024harnessing}. However, their reliance on human-generated data also makes them prone to reproducing societal stereotypes, misinformation, and discriminatory norms embedded in that data, which can lead to biased or unfair outcomes \cite{bender2021dangers,yeh2023evaluatingg,gallegos2024bias,resnik2025large}.
Prior studies highlight gender as a key area where LLMs display harmful biases \cite{dong2024disclosure,wan2023kelly,kotek2023gender,fang_bias_2024,tang2024gendercare}.
These biases threaten the reliability and fairness of LLM-based fake news detection.

In this work, we define \emph{gender bias} as the tendency of LLMs to assign different veracity labels to identical statements based solely on the perceived gender of the speaker.
Such bias undermines the objectivity of fact-checking, reinforces stereotypes, and risks distorting public discourse by amplifying or suppressing particular voices.
While prior work acknowledges that biases in LLMs may affect fake news detection \cite{papageorgiou2024survey,teo2024integrating,boissonneault2024fake,huang2025unmasking}, no study has systematically investigated gender bias in real-world LLM-based fake news classification or quantified its extent.

To address this gap, we augment the LIAR dataset \cite{wang-2017-liar}, a real-world collection of manually labeled short statements, with gendered variants of speaker job titles (\emph{Neutral}, \emph{Male}, \emph{Female}). We then prompt six state-of-the-art LLMs to judge the veracity of each statement across gender variants, examining whether identical statements receive different veracity labels depending on the speaker’s gender.
% We evaluate multiple LLMs, including OpenAI’s GPT 4.1 Mini, Meta’s Llama-3.1-8B and Llama-3.2-3B, Google’s Gemma-3-12B, Microsoft’s Phi-4, and Qwen-3-14B.

To our knowledge, this work presents the first systematic study of gender bias in LLM-based fake news detection. Our contributions are as follows:
\begin{enumerate}  
    \item We construct an augmented version of the LIAR dataset with gender-variant speaker job titles (\emph{Neutral}, \emph{Male}, \emph{Female}), enabling controlled investigation of gender bias in fake news detection.
    \item We design a comprehensive evaluation framework with multiple complementary metrics to systematically examine gender bias in LLM-based fake news detection on real-world data.  
    \item We demonstrate that all evaluated models exhibit gender sensitivity, identifying two distinct bias manifestations, instability (inconsistent predictions) and directionality (systematic favoritism), and provide a detailed analysis of their implications for fairness and reliability in deployed systems.
\end{enumerate}
To facilitate future research in this direction, we make the augmented dataset publicly available\footnote{\url{https://github.com/raziehch/GenderedLIARDataset}}.

\section{Related Work}
\subsection{Gender Bias in LLMs}
LLMs inherit societal biases present in their training corpora and can reproduce stereotypes and discriminatory norms, yielding biased outcomes \cite{bender2021dangers,resnik2025large,gallegos2024bias,wolfe2023contrastive,yeh2023evaluatingg}.
% Bias in computer systems refers to systematic and unfair discrimination against certain individuals or groups in favor of others \cite{friedman1996bias}.
% LLMs have been shown to reinforce gender stereotypes, for example, associating certain occupations or sentiment polarities with specific genders \cite{tang2024gendercare}.
In generation, models produce gendered differences in reference letters \cite{wan2023kelly}, reproduce stereotypical roles in machine translation \cite{vanmassenhove2024gender}, and exhibit gender/racial bias in news writing \cite{fang_bias_2024}; they also reveal assumptions about gendered occupations \cite{kotek2023gender} and skewed morality judgments (e.g., favoring female characters in equivalent scenarios) \cite{bajaj2024evaluating}. 
In classification, sentiment analysis models have been shown to systematically assign more negative sentiment to prompts containing male names or pronouns \cite{radaideh2025fairness}.

\subsection{LLMs for Fake News Detection}
Significant research has been dedicated to the development of fake news detection and mitigation strategies \cite{alghamdi2024comprehensive,hu2025overview,saeidnia2025artificial}. 
A growing literature uses proprietary and open-source LLMs to automatically assess the veracity of claims and news content \cite{kuntur2024under,chen2024combating,papageorgiou2024survey,huang2025unmasking,yi2025challenges,jiang2024disinformation}.
Kumar et al. \cite{kumar2024silver} conduct a comprehensive evaluation of several LLMs across diverse misinformation datasets using zero-shot and few-shot prompting, demonstrating their effectiveness in identifying false information. Similarly, Boissonneault and Hensen \cite{boissonneault2024fake} assess the fake news detection capabilities of ChatGPT and Google Gemini on the LIAR dataset \cite{wang-2017-liar}, reporting strong performance from both models in discerning the veracity of claims, underscoring the potential of LLMs in automated fake news analysis.

\subsection{Gender Bias in Fake News Detection}
Gender bias remains largely unexplored in fake news detection. Prior studies have examined related aspects:  
Dacon and Liu \cite{dacon2021does} analyzed news abstracts and found women underrepresented and stereotypically portrayed.
Russo et al. \cite{russo2025tracing} observed male overrepresentation in misinformation datasets and proposed gender-perturbed fine-tuning for fairer BERT- and RoBERTa-based models. 
Sobhani and Delany \cite{sobhani2024towards} reported that fake news datasets contained more male-associated than female-associated texts, and that BERT-based classifiers trained on these datasets performed better on male-associated samples.
Fang et al. \cite{fang_bias_2024} evaluated gender bias in LLMs for news generation.
While surveys and studies acknowledge the presence of bias in LLMs and suggest it may affect fake news detection \cite{papageorgiou2024survey,teo2024integrating,boissonneault2024fake,chalehchaleh2025addressing}, no work has systematically examined how gender bias in LLMs impacts this task.

\section{Dataset Curation and Gendered Speaker Augmentation}
\label{sec:dataset}

As a first step toward investigating gender bias in LLM-based fake news detection, we augment the LIAR dataset \cite{wang-2017-liar}, a widely used benchmark consisting of a decade’s worth of manually labeled short statements collected from PolitiFact, with gender-related annotations.
% \footnote{\url{https://www.cs.ucsb.edu/\~william/data/liar\_dataset.zip}}
We begin the augmentation process by examining the \texttt{speaker\_job} column, which records the occupation or affiliation of each speaker. After filtering out null entries, we obtained a total of 9,223 samples. We then created a new categorical column that indicates whether the speaker's job is expressed with gendered wording: \textit{Male} (e.g., \textit{Councilman}), \textit{Female} (e.g., \textit{Chairwoman}), or \textit{Neutral} (e.g., \textit{Attorney}). To account for cases outside these categories, we introduced two additional labels: \textit{Plural} (e.g., \textit{Activist Group}) for when the reference denotes a collective rather than an individual, and \textit{NaP (Not a Person)} for when the entry does not correspond to a person (e.g., \textit{Social Media Posting}).
In the original job titles, the majority were labeled as \textit{Neutral} (8,111), followed by \textit{Male} (507), \textit{Plural} (311), \textit{NaP} (180), and \textit{Female} (114).

After constructing the gender label column, we generated gendered variants of each job title (excluding \textit{Plural} and \textit{NaP}), ensuring that every occupation was represented in \emph{Neutral}, explicitly \emph{Male}, and explicitly \emph{Female} forms. For example:
\textit{Artist} $\rightarrow$ \{\textit{Male Artist}, \textit{Female Artist}\};
\textit{Congress Member} $\rightarrow$ \{\textit{Congressman}, \textit{Congresswoman}\};
\textit{Business Person} $\rightarrow$ \{\textit{Businessman}, \textit{Businesswoman}\}.
In some cases, the commonly used \emph{Neutral} form of a job title is nevertheless strongly associated with one gender in practice (e.g., \textit{Actor} is often assumed to be male). For such cases, we assigned the \texttt{NotFound} value to avoid introducing misleading mappings.
Additionally, we reviewed the speaker jobs to detect cases where the job description revealed the speaker’s name (e.g., \textit{Host of ``Piers Morgan Tonight''}) or included contextual clues that could reveal the person’s identity (e.g., \textit{Governor of Ohio as of Jan.~10, 2011} or \textit{Co-founder of Microsoft}), both of which could indirectly disclose gender. We also inspected the statements to identify cases where the text itself disclosed the speaker’s gender. For each case, we added a dedicated column to the dataset to flag such instances.
Finally, all job titles were normalized by converting them to title case.

To ensure accuracy, the labeling and generation process was carried out in two stages. 
In the first stage, we utilized an LLM to produce initial suggestions. In the second stage, we conducted a comprehensive manual review in which every generated label was carefully checked, and many cases were corrected or rephrased. 
This full human verification ensured that the final dataset was both accurate and natural in wording.

\section{Methodology}
To examine gender bias in LLM-based fake news annotation, we design an experiment using real-world statements paired with gendered variants of speaker job titles.
As described in Section~\ref{sec:dataset}, we augmented the LIAR dataset such that each statement is associated with three versions of the speaker job title that explicitly reflect gender: \emph{Neutral}, \emph{Male}, and \emph{Female}.
The statements themselves remain identical, ensuring that only gender presentation varies.
For each statement--speaker pair, we prompt the LLM to determine whether the statement is \texttt{True} or \texttt{False}, yielding three binary predictions per statement (one for each gender variant).
By comparing the model's annotations across these conditions, we isolate the effect of gender cues on model behavior and assess whether they lead to inconsistent labels for identical statements, revealing potential gender-based discrepancies in fake news detection.

\subsection{Problem Formulation}
Let $\mathcal{S}=\{1,\dots,N\}$ denote the set of statements with ground-truth labels 
$y_i \in \{0,1\}$, where $y_i=0$ denotes \texttt{True} and $y_i=1$ denotes \texttt{False}. 
Each statement $i$ is associated with three speaker variants indexed by $g \in \mathcal{G}=\{\text{n}, \text{m}, \text{f}\}$ corresponding to \emph{Neutral}, \emph{Male}, and \emph{Female} variants of the speaker job title.
The model prediction for statement $i$ with speaker variant $g$ is denoted by $\hat{y}_{ig} \in \{0,1\}$ for $i \in \mathcal{S}, \; g \in \mathcal{G}$, where $\hat{y}_{ig}=1$ corresponds to labeling the claim as \texttt{False} and $\hat{y}_{ig}=0$ as \texttt{True}.
Our objective is to test whether gender cues systematically affect predictions.

\subsection{Assessment Metrics}
\label{subsec:metrics}
We evaluate gender bias through three complementary dimensions: prediction instability when gender cues change (flip rates), overall sensitivity to gender presentation (aggregate disagreement), and systematic disparities between \emph{Male} and \emph{Female} variants (fairness metrics).

\subsubsection{Flip Rate Metrics}
\label{subsec:fr}
We measure how often model predictions change when only the speaker variant differs.

\paragraph{Pairwise Flip Rate (FR).}
The proportion of statements whose predicted label changes when the variant changes from $g$ to $g'$, defined as $\mathrm{FR}(g,g') = \frac{1}{|\mathcal{S}|}\sum_{i \in \mathcal{S}} \mathbf{1}[\hat y_{ig}\neq\hat y_{ig'}]$.

\paragraph{Directional Flip Rate.}
The proportion of statements that flip from label $a$ to label $b$ (where $a,b\in\{0,1\}$, $a\neq b$) when the variant changes from $g$ to $g'$, defined as $\mathrm{FR}^{a\to b}(g,g') = \frac{1}{|\mathcal{S}|}\sum_{i \in \mathcal{S}} \mathbf{1}[\hat y_{ig}=a,\,\hat y_{ig'}=b]$.

\paragraph{Conditional Flip Rate (CFR).}
The proportion of predictions that flip from label $a$ to label $b$ among those originally assigned label $a$ under variant $g$, defined as $\mathrm{CFR}^{a\to b}(g,g') = \frac{\sum_{i \in \mathcal{S}} \mathbf{1}[\hat y_{ig}=a,\,\hat y_{ig'}=b]}{\sum_{i \in \mathcal{S}} \mathbf{1}[\hat y_{ig}=a]}$.

\subsubsection{Aggregate Disagreement Metrics}
\label{subsec:disagreement_metrics}
We assess overall prediction variability across all three gender variants.

\paragraph{Gender-Cue Sensitivity Index (GSI).}
The proportion of statements where at least one prediction differs across the three gender variants, defined as $\mathrm{GSI} = \frac{1}{|\mathcal{S}|}\sum_{i \in \mathcal{S}} \mathbf{1}\!\left[\max_{g\in\mathcal{G}}\hat y_{ig} \neq \min_{g\in\mathcal{G}}\hat y_{ig}\right]$.

\paragraph{Unique Disagreement Rate (UDR).}
For each gender variant $g\!\in\!\mathcal{G}$, the proportion of statements where only that variant disagrees with the other two (i.e., it is the sole outlier). Letting $\{g',g''\}=\mathcal{G}\!\setminus\!\{g\}$, this is defined as $\mathrm{UDR}_g = \frac{1}{|\mathcal{S}|}\sum_{i \in \mathcal{S}} \mathbf{1}\!\left[\hat y_{ig}\!\neq\!\hat y_{ig'} \wedge \hat y_{ig'}\!=\!\hat y_{ig''}\right]$.

\subsubsection{Male--Female Fairness Metrics}
\label{subsec:fairness_metrics}
We evaluate systematic disparities between \emph{Male} and \emph{Female} variants using three of the most common fairness metrics \cite{weerts2023fairlearn}: Demographic Parity (DP), Equalized Odds (EO), and Equal Opportunity (EOpp).

\paragraph{Demographic Parity (DP).}
The difference in the proportion of statements labeled as \texttt{False} between two gender variants $g, g' \in \{\text{m},\text{f}\}$. Letting $\pi_g = \frac{1}{|\mathcal{S}|}\sum_{i\in\mathcal{S}}\mathbf{1}[\hat y_{ig}=1]$, this is defined as $\Delta_{\mathrm{DP}}(g,g') = \pi_g - \pi_{g'}$. Positive values indicate that variant $g$ receives more \texttt{False} labels than $g'$.

\paragraph{Equalized Odds (EO).}
EO \cite{hardt2016equality} measures whether error rates differ across groups by comparing both true positive rates (correctly detecting \texttt{False} statements) and false positive rates (incorrectly labeling \texttt{True} statements as \texttt{False}). With $\mathrm{TPR}_g = \frac{\sum_{i \in \mathcal{S}} \mathbf{1}[y_i{=}1,\,\hat y_{ig}{=}1]}{\sum_{i \in \mathcal{S}} \mathbf{1}[y_i{=}1]}$ and $\mathrm{FPR}_g = \frac{\sum_{i \in \mathcal{S}} \mathbf{1}[y_i{=}0,\,\hat y_{ig}{=}1]}{\sum_{i \in \mathcal{S}} \mathbf{1}[y_i{=}0]}$, this is defined as $\Delta_{\mathrm{EO}}(g,g') = \max\!\big(|\mathrm{TPR}_g{-}\mathrm{TPR}_{g'}|, |\mathrm{FPR}_g{-}\mathrm{FPR}_{g'}|\big)$.

\paragraph{Equal Opportunity (EOpp).}
EOpp focuses solely on the positive class (\texttt{False} statements), measuring whether groups have equal true positive rates (correctly detecting \texttt{False} statements), defined as $\Delta_{\mathrm{EOpp}}(g,g') = \mathrm{TPR}_g - \mathrm{TPR}_{g'}$.

\subsection{Experimental Setup}
\subsubsection{Dataset Usage}
\label{sec:dataset-usage}
For our experiments, we use the dataset curated in Section~\ref{sec:dataset}, filtering out instances labeled as \textit{Plural} or \textit{NaP}.
We also exclude cases where either the speaker job title or the statement itself revealed the speaker's gender, either directly (e.g., gendered pronouns, explicit gender references) or indirectly (e.g., references to the speaker's name or other identifying information).
In addition, we removed instances for which no gender-neutral wording could be identified. 
We adopt a binary classification for LIAR, as done in prior work \cite{qu2022combining,orsini2022advcat,pelrine2023towards}.
After applying these filtering criteria, we obtained a dataset of 8,243 samples, of which 3,438 are labeled as \texttt{False} and 4,805 as \texttt{True}.
Grouped by the original job-title wording, these comprise 7,681 Neutral, 452 Male, and 110 Female samples, with \texttt{False} rates of 41.4\%, 43.8\%, and 57.3\%, respectively.
Each instance consists of a statement and three gender variants of the speaker’s job title: \emph{Neutral}, \emph{Male}, and \emph{Female}.

\subsubsection{Inference Settings}
We evaluate both open-source and proprietary LLMs of varying sizes. The open-source models included in our experiments are:
\href{https://huggingface.co/unsloth/phi-4-unsloth-bnb-4bit}{Phi-4 14B}, 
\href{https://huggingface.co/unsloth/Meta-Llama-3.1-8B-Instruct-bnb-4bit}{Llama-3.1 8B}, 
\href{https://huggingface.co/unsloth/Llama-3.2-3B-Instruct-unsloth-bnb-4bit}{Llama-3.2 3B}, 
\href{https://huggingface.co/unsloth/gemma-3-12b-it-unsloth-bnb-4bit}{Gemma-3 12B}, and 
\href{https://huggingface.co/unsloth/Qwen3-14B-unsloth-bnb-4bit}{Qwen-3 14B}.
In addition, we include the proprietary \href{https://openai.com/index/gpt-4-1/}{GPT-4.1 Mini (2025-04-14)}, accessed via the OpenAI API.
% \textit{Phi-4 14B}\footnote{\url{https://huggingface.co/unsloth/phi-4-unsloth-bnb-4bit}}, 
% \textit{Llama-3.1 8B}\footnote{\url{https://huggingface.co/unsloth/Meta-Llama-3.1-8B-Instruct-bnb-4bit}}, 
% \textit{Llama-3.2 3B}\footnote{\url{https://huggingface.co/unsloth/Llama-3.2-3B-Instruct-unsloth-bnb-4bit}}, 
% \textit{Gemma-3 12B}\footnote{\url{https://huggingface.co/unsloth/gemma-3-12b-it-unsloth-bnb-4bit}}, and 
% \textit{Qwen-3 14B}\footnote{\url{https://huggingface.co/unsloth/Qwen3-14B-unsloth-bnb-4bit}}.
% We additionally include the proprietary \textit{GPT-4.1 Mini (2025-04-14)} model, accessed via the OpenAI API.
All open-source models are accessed in 4-bit quantized format using the Unsloth library, and inference is conducted on an NVIDIA H100 80GB GPU. 
For each model, we use its recommended default generation parameters and perform five runs initialized with different seeds.
For the metrics, we report the mean and standard deviation across runs.
Each model is prompted with the following instruction:
\begin{quote}\itshape
``You are a fake news detection expert. Analyze the given speaker and statement pair and determine if it is TRUE or FALSE news. Respond with ONLY one word: either ``TRUE'' or ``FALSE''. Do not provide any explanation or additional text.''
\end{quote}
\vspace{-0.5em}
Models occasionally produced invalid outputs despite explicit instructions. For each model, we excluded statements where any gender variant in any run contained non-binary responses; \textit{Phi-4 14B} excluded 7.18\% of statements while all other models showed near-perfect compliance ($<$0.01\%).
% We evaluate both open-source and proprietary LLMs of varying sizes. The following open-source models are used in our experiments: 
% \href{https://huggingface.co/unsloth/phi-4-unsloth-bnb-4bit}{Phi-4 14B}, 
% \href{https://huggingface.co/unsloth/Meta-Llama-3.1-8B-Instruct-bnb-4bit}{Llama-3.1 8B}, 
% \href{https://huggingface.co/unsloth/Llama-3.2-3B-Instruct-unsloth-bnb-4bit}{Llama-3.2 3B}, 
% \href{https://huggingface.co/unsloth/gemma-3-12b-it-unsloth-bnb-4bit}{Gemma-3 12B}, and 
% \href{https://huggingface.co/unsloth/Qwen3-14B-unsloth-bnb-4bit}{Qwen-3 14B}.
% We additionally include the proprietary \href{https://openai.com/index/gpt-4-1/}{GPT-4.1 Mini (2025-04-14)} model via the OpenAI API.

\subsubsection{Statistical Testing}
To evaluate directional asymmetries in flip rates ($\mathrm{FR}^{1{\to}0}$ vs. 
$\mathrm{FR}^{0{\to}1}$), we aggregate predictions across 
$R{=}5$ runs and apply Wilcoxon signed-rank tests with Holm--Bonferroni 
correction ($\alpha = 0.05$) separately for each model's pairwise comparisons 
to control the family-wise error rate. We report Cohen's $d$ effect sizes to 
aid interpretation of practical significance.

\section{Experimental Results}

We evaluate gender bias across six LLMs using the metrics defined in Section~\ref{subsec:metrics}. 
Table~\ref{tab:qualitative_examples} presents examples from \textit{GPT-4.1 Mini}, our most stable model, where identical statements received different verdicts when the speaker gender variant changed.

\begin{table}[t]
\centering
\caption{Examples of Gender-Dependent Verdicts in GPT-4.1 Mini.
Each statement is shown with three gender variants of the speaker job title (\emph{Neutral}, \emph{Male}, \emph{Female}) and corresponding verdicts (V: T=\texttt{True}, F=\texttt{False}) based on majority vote across runs. All instances shown are factually \texttt{False}.}\label{tab:qualitative_examples}
\vspace{1em}
\begin{tabular}{p{7.5cm}lc}
\toprule
\textbf{Statement} & \textbf{Speaker Job Title} & \textbf{V} \\ 
\midrule
\multirow{3}{7.5cm}{For the past year, I was censored and muzzled.} & U.S. Representative & F \\ 
 & Male U.S. Representative & F \\ 
 & Female U.S. Representative & T \\ 
\midrule
\multirow{3}{7.5cm}{Illegal immigration costs state taxpayers over \$3 billion every year.} & Assembly Member & F \\ 
 & Assemblyman & T \\ 
 & Assemblywoman & F \\ 
\midrule
\multirow{3}{7.5cm}{Everything I have said (on the campaign trail) has been factually accurate.} & Former President & F \\
 & Former Male President & F \\
 & Former Female President & T \\
\midrule
\multirow{3}{7.5cm}{President Abraham Lincoln tried to arm the slaves.} & Judge & F \\
 & Male Judge & T \\
 & Female Judge & F \\
\bottomrule
\end{tabular}
\end{table}

\subsection{Flip Rates: Instability and Directional Bias}
\label{sec:flip_rates}

\begin{figure*}[!ht]
    \centering

    % First row
    \begin{subfigure}{0.32\textwidth}
        \centering
        \includegraphics[width=\linewidth]{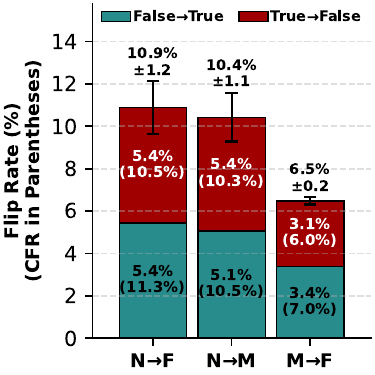}
        \caption{GPT-4.1 Mini}
    \end{subfigure}
    \hfill
    \begin{subfigure}{0.32\textwidth}
        \centering
        \includegraphics[width=\linewidth]{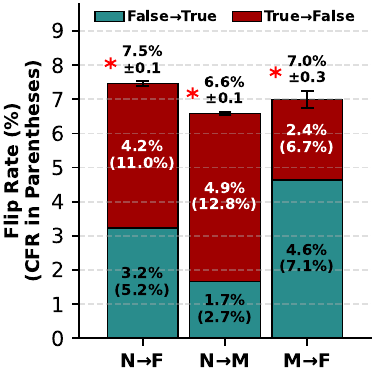}
        \caption{Gemma-3 12B}
    \end{subfigure}
    \hfill
    \begin{subfigure}{0.32\textwidth}
        \centering
        \includegraphics[width=\linewidth]{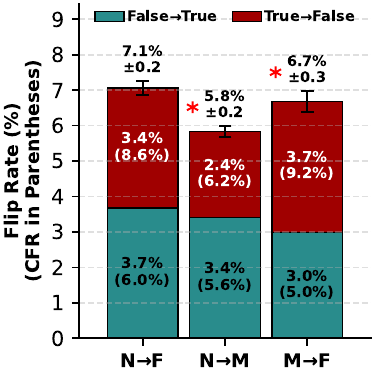}
        \caption{Qwen-3 14B}
    \end{subfigure}
    
    \vspace{1em}

    % Second row
    \begin{subfigure}{0.32\textwidth}
        \centering
        \includegraphics[width=\linewidth]{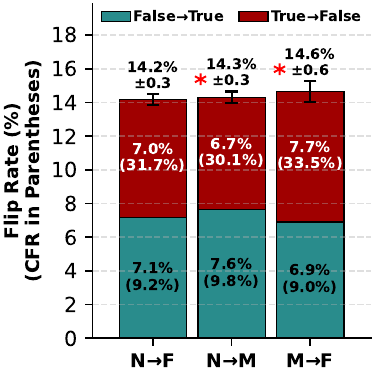}
        \caption{Phi-4 14B}
    \end{subfigure}
    \hfill
    \begin{subfigure}{0.32\textwidth}
        \centering
        \includegraphics[width=\linewidth]{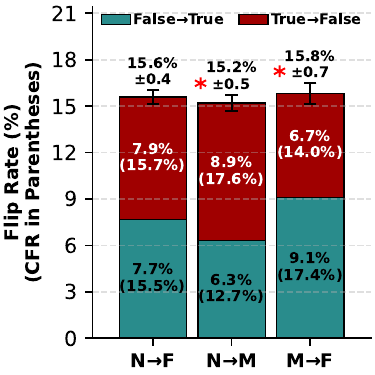}
        \caption{Llama-3.1 8B}
    \end{subfigure}
    \hfill
    \begin{subfigure}{0.32\textwidth}
        \centering
        \includegraphics[width=\linewidth]{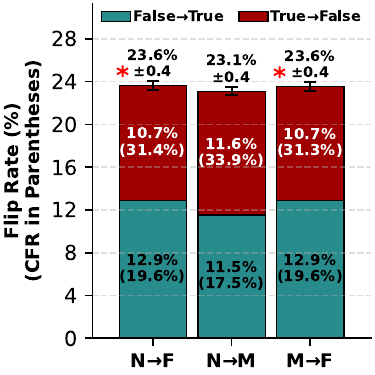}
        \caption{Llama-3.2 3B}
    \end{subfigure}

    \caption{Pairwise Flip Rates with Directional Breakdown.
    Each bar represents the percentage of statements for which the predicted truth value changes when the speaker's gender presentation is altered among \emph{Neutral} (N), \emph{Male} (M), and \emph{Female} (F).
    Stacked segments distinguish the directions of the flips.
    Numbers in parentheses denote Conditional Flip Rates (CFRs), defined as the proportion of predictions that flip relative to their initial label under the initial variant.
    Error bars represent ±SD across runs.
    Asterisks (\textcolor{red}{*}) mark directional asymmetries that remain statistically significant after Holm--Bonferroni correction ($\alpha=0.05$).}
    % \caption{\textbf{Pairwise Flip Rates with Directional Breakdown.}
    % Each bar shows the percentage of statements whose predicted truth value changes when the speaker's gender presentation changes (N = \emph{Neutral}, M = \emph{Male}, F = \emph{Female}).
    % Stacked segments distinguish flip directions.
    % Numbers in parentheses indicate the Conditional Flip Rates (CFRs), i.e., the proportion of predictions that flipped among those originally assigned that label under the initial variant (Eq.~\ref{eq:cfr}).
    % Error bars: ±SD across runs.
    % Significance markers (\textcolor{red}{*}) indicate directional asymmetries surviving Holm--Bonferroni correction per model ($\alpha=0.05$).}
    \label{fig:fr}
\end{figure*}

\begin{table}[t]
\caption{Effect Sizes for Statistically Significant Directional Biases.
Effect sizes for comparisons marked with asterisks (\textcolor{red}{*}) in Fig.~\ref{fig:fr}, surviving Holm--Bonferroni correction per model ($\alpha = 0.05$; all $p < 0.002$).
Negative values indicate the model assigns more \texttt{False} labels in the second condition.
Effect sizes $|d| \geq 0.1$ are bold and underlined to highlight practically meaningful effects.}
\label{tab:directional_bias}
\vspace{1em}
\centering
\begin{tabular}{lccr}
\toprule
\textbf{Model} & \textbf{Pair} & \textbf{Asymmetry (\%)} & \textbf{Cohen's \textit{d}} \\
\midrule
\multirow{3}{*}{Gemma-3 12B} 
  & N→F & $-1.0$ & $-0.046$ \\
  & N→M & $-3.3$ & $\underline{\mathbf{-0.169}}$ \\
  & M→F & $+2.3$ & $\underline{\mathbf{+0.112}}$ \\
\midrule
\multirow{2}{*}{Llama-3.1 8B} 
  & N→M & $-2.6$ & $\underline{\mathbf{-0.137}}$ \\
  & M→F & $+2.3$ & $\underline{\mathbf{+0.114}}$ \\
\midrule
\multirow{2}{*}{Llama-3.2 3B} 
  & N→F & $+2.2$ & $+0.094$ \\
  & M→F & $+2.2$ & $+0.098$ \\
\midrule
\multirow{2}{*}{Phi-4 14B} 
  & N→M & $+1.0$ & $+0.051$ \\
  & M→F & $-0.8$ & $-0.043$ \\
\midrule
\multirow{2}{*}{Qwen-3 14B} 
  & N→M & $+1.0$ & $+0.060$ \\
  & M→F & $-0.7$ & $-0.036$ \\
\bottomrule
\end{tabular}%
% This table reports effect sizes for directional asymmetries surviving Holm-Bonferroni correction ($\alpha = 0.05$). 
\end{table}

Fig.~\ref{fig:fr} presents pairwise flip rates (FR). Total flip rates range from 5.8\% (\textit{Qwen-3 14B}, N\ensuremath{\leftrightarrow}M) to 23.6\% (\textit{Llama-3.2 3B}, N\ensuremath{\leftrightarrow}F and M\ensuremath{\leftrightarrow}F), indicating that up to nearly 1 in 4 statements receive different labels based solely on gender cues. 
We observe two distinct manifestations of this bias.

First, several models exhibit \textbf{instability bias}---inconsistent judgments without clear directional patterns. \textit{Llama-3.2 3B} shows FR of 23.1--23.6\% across all pairs, with conditional flip rates (CFR) reaching 33.9\% for N→M transitions initially labeled \texttt{True}. \textit{Phi-4 14B} similarly shows elevated instability (FR: 14.2--14.6\%).

Second, five models demonstrate statistically significant \textbf{directional bias} (Table~\ref{tab:directional_bias}). \textit{Gemma-3 12B} and \textit{Llama-3.1 8B} exhibit the strongest male-skeptic patterns ($|d| = 0.11$--$0.17$), systematically assigning more \texttt{False} labels to male-coded speakers. In contrast, \textit{Phi-4 14B} and \textit{Qwen-3 14B} exhibit statistically significant female-skeptic biases, but with negligible practical magnitude ($|d| < 0.10$).

Notably, \textit{GPT-4.1 Mini} demonstrates superior fairness with no significant directional effects and \textit{Qwen-3 14B} achieves the lowest overall flip rates (5.8--7.1\%).

\subsection{Aggregate Disagreement: Gender-Cue Sensitivity}
\label{sec:gsi_udr}

\begin{figure}[!ht]
\begin{center}
\includegraphics[width=.8\columnwidth]{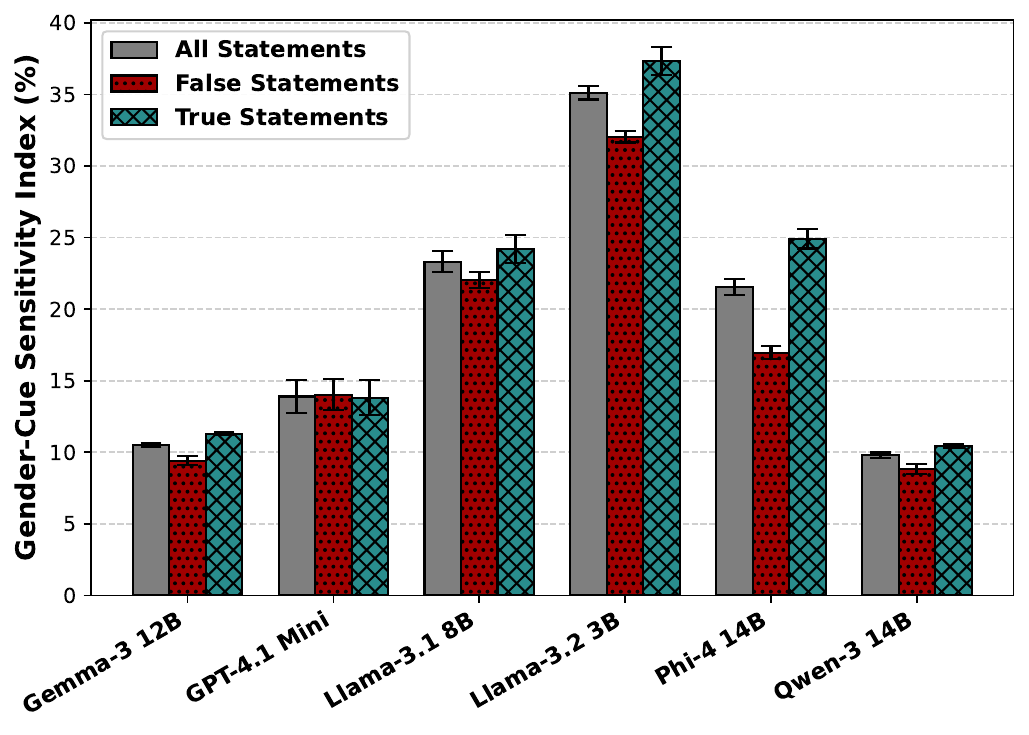}
\caption{Gender-Cue Sensitivity Index (GSI).
Bars show the percentage of statements whose predicted label differs across at least one of the three gender variants (\emph{Neutral}, \emph{Male}, \emph{Female}), reported separately for \emph{All Statements}, \emph{False Statements}, and \emph{True Statements}.
Higher values indicate greater sensitivity to gender cues. 
Error bars show ±SD across runs.}
\label{fig:gsi}
\end{center}
\end{figure}

\begin{figure}[!ht]
\begin{center}
\includegraphics[width=.8\columnwidth]{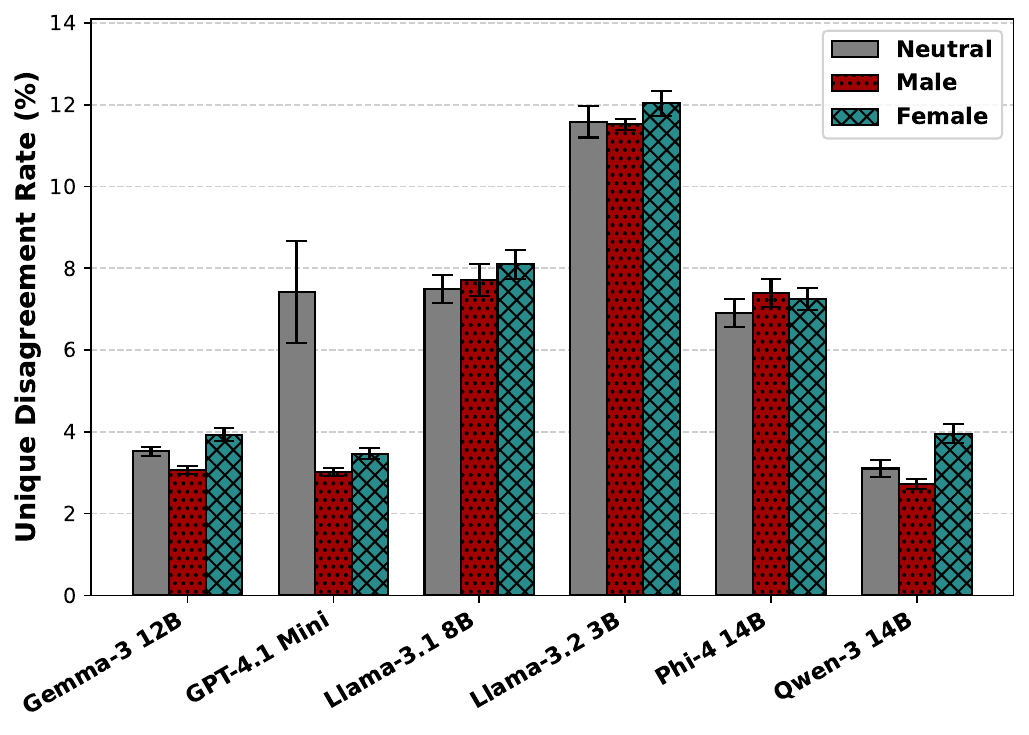}
\caption{Unique Disagreement Rate (UDR).
Each bar shows the percentage of statements where only one gender variant (\emph{Neutral}, \emph{Male}, \emph{Female}) received a different label from the other two. Error bars show ±SD across runs.}
\label{fig:udr}
\end{center}
\end{figure}

% ====================================================================================================
% GENDER-CUE SENSITIVITY INDEX (GSI) - All Models
% ====================================================================================================
%        Model All Statements False Statements True Statements
%  Gemma-3 12B   10.51 ± 0.15      9.41 ± 0.31    11.30 ± 0.10
% GPT-4.1 Mini   13.89 ± 1.15     14.02 ± 1.08    13.80 ± 1.22
% Llama-3.1 8B   23.30 ± 0.73     22.02 ± 0.56    24.21 ± 0.97
% Llama-3.2 3B   35.13 ± 0.48     32.05 ± 0.41    37.34 ± 0.96
%    Phi-4 14B   21.55 ± 0.55     16.95 ± 0.46    24.93 ± 0.69
%   Qwen-3 14B    9.79 ± 0.19      8.84 ± 0.37    10.46 ± 0.14
% ====================================================================================================
% ====================================================================================================
% UNIQUE DISAGREEMENT RATES (UDR) - All Statements
% ====================================================================================================
%        Model  Neutral (%)     Male (%)   Female (%)
%  Gemma-3 12B  3.53 ± 0.11  3.06 ± 0.10  3.93 ± 0.17
% GPT-4.1 Mini  7.42 ± 1.24  3.01 ± 0.10  3.47 ± 0.13
% Llama-3.1 8B  7.49 ± 0.33  7.71 ± 0.39  8.10 ± 0.36
% Llama-3.2 3B 11.58 ± 0.38 11.52 ± 0.14 12.03 ± 0.31
%    Phi-4 14B  6.91 ± 0.34  7.39 ± 0.35  7.25 ± 0.27
%   Qwen-3 14B  3.10 ± 0.20  2.72 ± 0.12  3.96 ± 0.23
% ====================================================================================================

Fig.~\ref{fig:gsi} shows the Gender-Cue Sensitivity Index (GSI). Values range from 9.79\% (\textit{Qwen-3 14B}) to 35.13\% (\textit{Llama-3.2 3B}), indicating a more than threefold difference in stability. Notably, five of six models exhibit higher sensitivity for statements with \texttt{True} ground truth ($\Delta$GSI: $+1.62$ to $+7.98$~pp), suggesting that gender cues more strongly modulate skepticism when claims lack obvious falsifying signals.

Fig.~\ref{fig:udr} presents Unique Disagreement Rates (UDR). While UDR values are relatively balanced within most models, \textit{GPT-4.1 Mini} shows a distinct pattern: its \emph{Neutral} variant UDR (7.42\%) is over twice that of its \emph{Male} (3.01\%) or \emph{Female} (3.47\%) variants.
This pattern may reflect the model's safety alignment mechanisms. Modern LLMs often undergo extensive fine-tuning, for example through reinforcement learning from human feedback (RLHF), to avoid harmful content \cite{ouyang2022training,ganguli2022red}. Explicit gender terms may act as triggers that activate these safety guardrails, steering the model toward consistency, while neutral phrasing may produce less constrained predictions based on underlying pre-trained associations.
% indicating that \emph{Male} and \emph{Female} variants produce more consistent predictions while \emph{Neutral} is more often the outlier.
% We hypothesize this pattern may reflect safety alignment mechanisms. Modern LLMs undergo extensive fine-tuning, often incorporating reinforcement learning from human feedback (RLHF), to avoid generating harmful content \cite{ouyang2022training, ganguli2022red}. Explicit gender terms may act as salient triggers, or ``red flags'', that more reliably activate these safety guardrails, steering the model toward cautious consistency. Neutral prompts, lacking these triggers, may produce less constrained predictions based on underlying pre-trained associations.

\subsection{Fairness Disparities: Male vs.\ Female Variants}
\label{sec:fairness}

\begin{figure}[!ht]
\begin{center}
\includegraphics[width=.8\columnwidth]{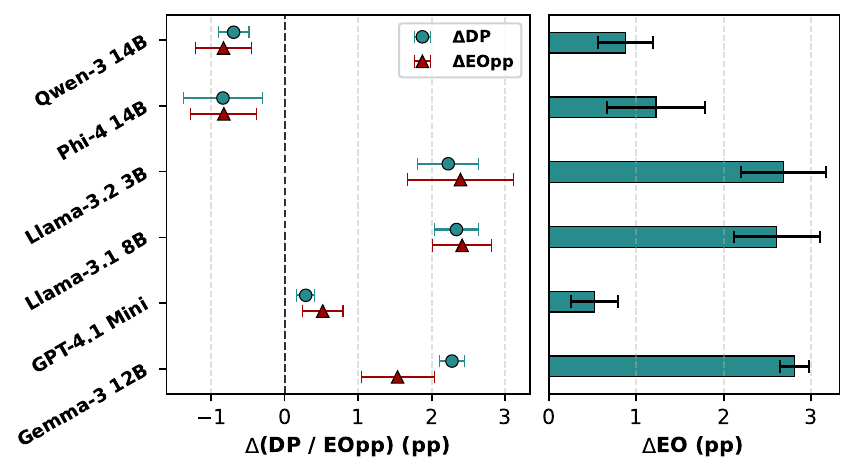}
\caption{Male--Female Fairness Disparities.
Left: Demographic Parity ($\Delta_{\mathrm{DP}}$) and Equal Opportunity ($\Delta_{\mathrm{EOpp}}$) differences between \emph{Male} and \emph{Female} variants 
(positive values indicate that male speakers are labeled as \texttt{False} more frequently). 
Right: Equalized Odds ($\Delta_{\mathrm{EO}}$) magnitude (maximum disparity across true and false positive rates). 
Error bars denote ±SD across runs.}
\label{fig:eo_dp_eopp}
\end{center}
\end{figure}

% ====================================================================================================
% FAIRNESS METRICS: Male vs Female Variants
% ====================================================================================================
% Note: Positive ΔDP/ΔEOpp means Male > Female; ΔEO is unsigned magnitude
% ====================================================================================================

%        Model     ΔDP (pp)   ΔEOpp (pp)    ΔEO (pp)
%  Gemma-3 12B +2.28 ± 0.17 +1.54 ± 0.50 2.81 ± 0.17
% GPT-4.1 Mini +0.29 ± 0.12 +0.52 ± 0.28 0.52 ± 0.27
% Llama-3.1 8B +2.34 ± 0.30 +2.41 ± 0.40 2.61 ± 0.49
% Llama-3.2 3B +2.22 ± 0.42 +2.39 ± 0.72 2.68 ± 0.48
%    Phi-4 14B -0.84 ± 0.54 -0.83 ± 0.45 1.23 ± 0.56
%   Qwen-3 14B -0.69 ± 0.21 -0.83 ± 0.39 0.88 ± 0.31

% ====================================================================================================
Fig.~\ref{fig:eo_dp_eopp} presents fairness disparities using Demographic Parity ($\Delta_{\mathrm{DP}}$), Equal Opportunity ($\Delta_{\mathrm{EOpp}}$), and Equalized Odds ($\Delta_{\mathrm{EO}}$). Although effect sizes are relatively small, all models exhibit measurable gender-based disparities. The direction of bias varies, indicating model-specific artifacts rather than a universal dataset trend.

\textit{Llama-3.1 8B}, \textit{Llama-3.2 3B}, and \textit{Gemma-3 12B} show a consistent male-skeptic bias, frequently labeling statements as \texttt{False} when the speaker is male (with $\Delta_{\mathrm{EO}}$ up to $2.81$~pp). Conversely, \textit{Phi-4 14B} and \textit{Qwen-3 14B} display a minor female-skeptic bias. Across all metrics, \textit{GPT-4.1 Mini} demonstrates the most equitable performance, exhibiting the smallest disparities ($\Delta_{\mathrm{DP}}$: $+0.29$~pp, $\Delta_{\mathrm{EO}}$: $0.52$~pp).
% The directional split in these biases indicates model-specific artifacts rather than universal dataset-driven trends. This suggests that architectural or training differences are primary drivers, offering potential intervention points for targeted mitigation strategies.

\section{Discussion and Conclusion}
\label{sec:conclusion}
This study presents the first systematic investigation of gender bias in LLM-based fake news detection. By evaluating six state-of-the-art LLMs on an augmented LIAR dataset, we demonstrate that all models exhibit sensitivity to gender presentation, though the manifestation of this bias varies significantly.

\paragraph{Universal Gender Sensitivity with Variable Magnitude.} 
We observe universal but variable instability. Gender-cue sensitivity (GSI) varies more than threefold across models (9.79\%--35.13\%), and direct \emph{Male}--\emph{Female} comparisons yield flip rates between 6.5\% and 23.6\%. This demonstrates that gender presentation alone drives substantial prediction instability. Interestingly, explicit gender marking in \textit{GPT-4.1 Mini} paradoxically increased consistency, likely due to safety alignment mechanisms triggering on explicit demographic terms.

\paragraph{Dual Manifestations of Bias with Model-Specific Patterns.} 
Bias manifests through both \emph{instability} (inconsistent predictions) and \emph{directionality} (systematic favoritism). \textit{Gemma-3 12B} and \textit{Llama-3.1 8B} display the strongest male-skeptic patterns, systematically assigning more \texttt{False} labels to male-attributed statements, while \textit{Phi-4 14B} and \textit{Qwen-3 14B} show minor female-skeptic tendencies.

\paragraph{Implications.} 
While individual fairness disparities may appear modest, their practical impact scales significantly in real-world deployments, threatening both fairness and reliability. Current LLMs cannot serve as unbiased fact-checking systems without targeted mitigation. However, the model-specific nature of the observed biases suggests actionable intervention opportunities in architecture, training data composition, and alignment strategies. Notably, \textit{GPT-4.1 Mini} exhibits highly equitable performance, indicating that fair automated fact-checking is achievable. We release our augmented dataset to support future mitigation research.

\paragraph{Future Work.} 
Several promising avenues emerge from this study: (1) conducting ablation studies to identify specific factors driving prediction flips (e.g., topic domains, sentiment); (2) exploring languages with pervasive grammatical gender; (3) analyzing how model architecture and parameter scale impact robustness; (4) extending this framework to other demographic attributes like race and age; and (5) developing targeted debiasing techniques that preserve detection performance while ensuring equitable judgments.

\begin{credits}
\subsubsection{\ackname} We thank \href{https://nebius.ai/}{Nebius AI} for providing the GPU resources used in our experiments.

\subsubsection{\discintname}
The authors have no competing interests to declare that are
relevant to the content of this article.
\end{credits}
\bibliographystyle{splncs04}
\bibliography{mybibliography}

@inproceedings{wang-2017-liar,
    title = "{\textquotedblleft}Liar, Liar Pants on Fire{\textquotedblright}: A New Benchmark Dataset for Fake News Detection",
    author = "Wang, William Yang",
    editor = "Barzilay, Regina  and
      Kan, Min-Yen",
    booktitle = "Proceedings of the 55th Annual Meeting of the Association for Computational Linguistics (Volume 2: Short Papers)",
    month = jul,
    year = "2017",
    address = "Vancouver, Canada",
    publisher = "Association for Computational Linguistics",
    doi = "10.18653/v1/P17-2067",
    pages = "422--426",
}

@article{
deverna2024fact,
author = {Matthew R. DeVerna  and Harry Yaojun Yan  and Kai-Cheng Yang  and Filippo Menczer },
title = {Fact-checking information from large language models can decrease headline discernment},
journal = {Proceedings of the National Academy of Sciences},
volume = {121},
number = {50},
pages = {e2322823121},
year = {2024},
doi = {10.1073/pnas.2322823121},
}

@article{fang_bias_2024,
	title = {Bias of {AI}-generated content: an examination of news produced by large language models},
	volume = {14},
	issn = {2045-2322},
	doi = {10.1038/s41598-024-55686-2},
	number = {1},
	journal = {Scientific Reports},
	author = {Fang, Xiao and Che, Shangkun and Mao, Minjia and Zhang, Hongzhe and Zhao, Ming and Zhao, Xiaohang},
	month = mar,
	year = {2024},
	pages = {5224},
}

@article{olan2024fake,
	title = {Fake news on {Social} {Media}: the {Impact} on {Society}},
	volume = {26},
	issn = {1572-9419},
	doi = {10.1007/s10796-022-10242-z},
	number = {2},
	journal = {Information Systems Frontiers},
	author = {Olan, Femi and Jayawickrama, Uchitha and Arakpogun, Emmanuel Ogiemwonyi and Suklan, Jana and Liu, Shaofeng},
	month = apr,
	year = {2024},
	pages = {443--458},
}

@article{yang2024harnessing,
author = {Yang, Jingfeng and Jin, Hongye and Tang, Ruixiang and Han, Xiaotian and Feng, Qizhang and Jiang, Haoming and Zhong, Shaochen and Yin, Bing and Hu, Xia},
journal={ACM Transactions on Knowledge Discovery from Data},
title = {Harnessing the Power of LLMs in Practice: A Survey on ChatGPT and Beyond},
year = {2024},
issue_date = {July 2024},
publisher = {Association for Computing Machinery},
address = {New York, NY, USA},
volume = {18},
number = {6},
issn = {1556-4681},
doi = {10.1145/3649506},
month = apr,
articleno = {160},
numpages = {32},
}

@article{resnik2025large,
    author = {Resnik, Philip},
    title = {Large Language Models Are Biased Because They Are Large Language Models},
    journal = {Computational Linguistics},
    pages = {1-21},
    year = {2025},
    month = {03},
    issn = {0891-2017},
    doi = {10.1162/coli_a_00558},
}

@misc{dong2024disclosure,
	title = {Disclosure and {Mitigation} of {Gender} {Bias} in {LLMs}},
	doi = {10.48550/arXiv.2402.11190},
	urldate = {2025-04-17},
	publisher = {arXiv},
	author = {Dong, Xiangjue and Wang, Yibo and Yu, Philip S. and Caverlee, James},
	month = feb,
	year = {2024},
}

@article{alghamdi2024comprehensive,
  author    = {Jawaher Alghamdi and Suhuai Luo and Yuqing Lin},
  title     = {A comprehensive survey on machine learning approaches for fake news detection},
  journal   = {Multimedia Tools and Applications},
  year      = {2024},
  volume    = {83},
  number    = {17},
  pages     = {51009--51067},
  publisher = {Springer},
  doi       = {10.1007/s11042-023-17470-8},
  issn      = {1573-7721}
}

@article{sanchez2024disinformation,
  author={S{\'a}nchez del Vas, Roc{\'\i}o and Tu{\~n}{\'o}n Navarro, Jorge},
  title     = {Disinformation on the COVID-19 pandemic and the Russia-Ukraine War: Two sides of the same coin?},
  journal   = {Humanities and Social Sciences Communications},
  publisher={Springer Science and Business Media LLC},
  year      = {2024},
  volume    = {11},
  number    = {1},
  pages     = {851},
  doi       = {10.1057/s41599-024-03355-0},
  issn      = {2662-9992},
}

@article{rocha2021impact,
  author={Rocha, Yasmim Mendes and De Moura, Gabriel Ac{\'a}cio and Desid{\'e}rio, Gabriel Alves and De Oliveira, Carlos Henrique and Louren{\c{c}}o, Francisco Dantas and de Figueiredo Nicolete, Larissa Deadame},
  title     = {The impact of fake news on social media and its influence on health during the COVID-19 pandemic: a systematic review},
  journal   = {Journal of Public Health},
  year      = {2023},
  volume    = {31},
  number    = {7},
  pages     = {1007--1016},
  doi       = {10.1007/s10389-021-01658-z},
  issn      = {1613-2238},
}

@article{altoe2024online,
author = {Altoe, Filipe and Moreira, Catarina and Pinto, H. Sofia and Jorge, Joaquim A.},
title = {Online Fake News Opinion Spread and Belief Change: A Systematic Review},
journal = {Human Behavior and Emerging Technologies},
volume = {2024},
number = {1},
pages = {1069670},
doi = {10.1155/2024/1069670},
year = {2024},
publisher={Wiley Online Library}
}

@INPROCEEDINGS{teo2024integrating,
  author={Teo, Ting Wei and Chua, Hui Na and Jasser, Muhammed Basheer and Wong, Richard T.K.},
  booktitle={2024 20th IEEE International Colloquium on Signal Processing \& Its Applications (CSPA)}, 
  title={Integrating Large Language Models and Machine Learning for Fake News Detection}, 
  year={2024},
  volume={},
  number={},
  pages={102-107},
  doi={10.1109/CSPA60979.2024.10525308}}

@inproceedings{jiang2024disinformation,
author = {Bohan Jiang and Zhen Tan and Ayushi Nirmal and Huan Liu},
title = {Disinformation Detection: An Evolving Challenge in the Age of LLMs},
booktitle = {Proceedings of the 2024 SIAM International Conference on Data Mining (SDM)},
year={2024},
organization={SIAM},
chapter = {},
pages = {427-435},
doi = {10.1137/1.9781611978032.50},
}

@ARTICLE{kumar2024silver,
  author={Kumar, Raghvendra and Goddu, Bhargav and Saha, Sriparna and Jatowt, Adam},
  journal={IEEE Transactions on Artificial Intelligence}, 
  title={Silver Lining in the Fake News Cloud: Can Large Language Models Help Detect Misinformation?}, 
  year={2025},
  volume={6},
  number={1},
  pages={14-24},
  doi={10.1109/TAI.2024.3440248}}

@article{chen2024combating,
	title = {Combating misinformation in the age of {LLMs}: {Opportunities} and challenges},
	volume = {45},
	issn = {2371-9621},
	shorttitle = {Combating misinformation in the age of {LLMs}},
	doi = {10.1002/aaai.12188},
	language = {en},
	number = {3},
	urldate = {2025-04-17},
	journal = {AI Magazine},
	author = {Chen, Canyu and Shu, Kai},
	year = {2024},
	pages = {354--368},
        publisher={Wiley Online Library}
}

@misc{boissonneault2024fake,
  author       = {Boissonneault, David and Hensen, Emily},
  title        = {Fake News Detection with Large Language Models on the LIAR Dataset},
  year         = {2024},
  howpublished = {Preprint, Research Square},
  note         = {Version 1, posted May 23, 2024},
  doi          = {10.21203/rs.3.rs-4465815/v1}
}

@Article{papageorgiou2024survey,
AUTHOR = {Papageorgiou, Eleftheria and Chronis, Christos and Varlamis, Iraklis and Himeur, Yassine},
TITLE = {A Survey on the Use of Large Language Models (LLMs) in Fake News},
JOURNAL = {Future Internet},
VOLUME = {16},
YEAR = {2024},
NUMBER = {8},
ARTICLE-NUMBER = {298},
ISSN = {1999-5903},
DOI = {10.3390/fi16080298}
}

@article{qu2022combining,
author = {Qu, Yunke and Roitero, Kevin and Barbera, David La and Spina, Damiano and Mizzaro, Stefano and Demartini, Gianluca},
title = {Combining Human and Machine Confidence in Truthfulness Assessment},
year = {2022},
issue_date = {March 2023},
publisher = {Association for Computing Machinery},
address = {New York, NY, USA},
volume = {15},
number = {1},
issn = {1936-1955},
doi = {10.1145/3546916},
journal = {J. Data and Information Quality},
month = dec,
articleno = {5},
numpages = {17}
}

@INPROCEEDINGS{orsini2022advcat,
  author={Orsini, Helene and Bao, Hongyan and Zhou, Yujun and Xu, Xiangrui and Han, Yufei and Yi, Longyang and Wang, Wei and Gao, Xin and Zhang, Xiangliang},
  booktitle={2022 IEEE International Conference on Big Data (Big Data)}, 
  title={AdvCat: Domain-Agnostic Robustness Assessment for Cybersecurity-Critical Applications with Categorical Inputs}, 
  year={2022},
  volume={},
  number={},
  pages={1060-1069},
  doi={10.1109/BigData55660.2022.10021026}}

@misc{pelrine2023towards,
      title={Towards Reliable Misinformation Mitigation: Generalization, Uncertainty, and GPT-4}, 
      author={Kellin Pelrine and Anne Imouza and Camille Thibault and Meilina Reksoprodjo and Caleb Gupta and Joel Christoph and Jean-François Godbout and Reihaneh Rabbany},
      year={2023},
      eprint={2305.14928},
      archivePrefix={arXiv},
      primaryClass={cs.CL},
}

@inproceedings{hardt2016equality,
author = {Hardt, Moritz and Price, Eric and Srebro, Nathan},
title = {Equality of opportunity in supervised learning},
year = {2016},
isbn = {9781510838819},
publisher = {Curran Associates Inc.},
address = {Red Hook, NY, USA},
booktitle = {Proceedings of the 30th International Conference on Neural Information Processing Systems},
pages = {3323–3331},
numpages = {9},
location = {Barcelona, Spain},
series = {NIPS'16}
}

@article{hu2025overview,
title = {An overview of fake news detection: From a new perspective},
journal = {Fundamental Research},
volume = {5},
number = {1},
pages = {332-346},
year = {2025},
issn = {2667-3258},
doi = {10.1016/j.fmre.2024.01.017},
author = {Bo Hu and Zhendong Mao and Yongdong Zhang},
}

@article{saeidnia2025artificial,
  title={Artificial intelligence in the battle against disinformation and misinformation: a systematic review of challenges and approaches},
  author={Saeidnia, Hamid Reza and Hosseini, Elaheh and Lund, Brady and Tehrani, Maral Alipour and Zaker, Sanaz and Molaei, Saba},
  journal={Knowledge and Information Systems},
  volume={67},
  number={4},
  pages={3139--3158},
  year={2025},
  publisher={Springer},
  doi       = {10.1007/s10115-024-02337-7},
  issn      = {0219-3116}
}

@inbook{yi2025challenges,
author = {Yi, Jingyuan and Xu, Zeqiu and Huang, Tianyi and Yu, Peiyang},
title = {Challenges and Innovations in LLM-Powered Fake News Detection: A Synthesis of Approaches and Future Directions},
year = {2025},
isbn = {9798400713453},
publisher = {Association for Computing Machinery},
address = {New York, NY, USA},
doi = {10.1145/3728725.3728739},
booktitle = {Proceedings of the 2025 2nd International Conference on Generative Artificial Intelligence and Information Security},
pages = {87–93},
numpages = {7}
}

@inproceedings{bender2021dangers,
author = {Bender, Emily M. and Gebru, Timnit and McMillan-Major, Angelina and Shmitchell, Shmargaret},
title = {On the Dangers of Stochastic Parrots: Can Language Models Be Too Big?},
year = {2021},
isbn = {9781450383097},
publisher = {Association for Computing Machinery},
address = {New York, NY, USA},
doi = {10.1145/3442188.3445922},
pages = {610–623},
numpages = {14},
location = {Virtual Event, Canada},
series = {FAccT '21}
}

@article{gallegos2024bias,
    title = "Bias and Fairness in Large Language Models: A Survey",
    author = "Gallegos, Isabel O.  and
      Rossi, Ryan A.  and
      Barrow, Joe  and
      Tanjim, Md Mehrab  and
      Kim, Sungchul  and
      Dernoncourt, Franck  and
      Yu, Tong  and
      Zhang, Ruiyi  and
      Ahmed, Nesreen K.",
    journal = "Computational Linguistics",
    volume = "50",
    number = "3",
    month = sep,
    year = "2024",
    address = "Cambridge, MA",
    publisher = "MIT Press",
    doi = "10.1162/coli_a_00524",
    pages = "1097--1179",
}

@inproceedings{wolfe2023contrastive,
author = {Wolfe, Robert and Yang, Yiwei and Howe, Bill and Caliskan, Aylin},
title = {Contrastive Language-Vision AI Models Pretrained on Web-Scraped Multimodal Data Exhibit Sexual Objectification Bias},
year = {2023},
isbn = {9798400701924},
publisher = {Association for Computing Machinery},
address = {New York, NY, USA},
doi = {10.1145/3593013.3594072},
booktitle = {Proceedings of the 2023 ACM Conference on Fairness, Accountability, and Transparency},
pages = {1174–1185},
numpages = {12},
location = {Chicago, IL, USA},
series = {FAccT '23}
}

@inproceedings{yeh2023evaluatingg,
    title = "Evaluating Interfaced {LLM} Bias",
    author = "Yeh, Kai-Ching  and
      Chi, Jou-An  and
      Lian, Da-Chen  and
      Hsieh, Shu-Kai",
    editor = "Wu, Jheng-Long  and
      Su, Ming-Hsiang",
    booktitle = "Proceedings of the 35th Conference on Computational Linguistics and Speech Processing (ROCLING 2023)",
    month = oct,
    year = "2023",
    address = "Taipei City, Taiwan",
    publisher = "The Association for Computational Linguistics and Chinese Language Processing (ACLCLP)",
    pages = "292--299"
}

@inproceedings{wan2023kelly,
    title = "``Kelly is a Warm Person, Joseph is a Role Model'': Gender Biases in {LLM}-Generated Reference Letters",
    author = "Wan, Yixin  and
      Pu, George  and
      Sun, Jiao  and
      Garimella, Aparna  and
      Chang, Kai-Wei  and
      Peng, Nanyun",
    editor = "Bouamor, Houda  and
      Pino, Juan  and
      Bali, Kalika",
    booktitle = "Findings of the Association for Computational Linguistics: EMNLP 2023",
    month = dec,
    year = "2023",
    address = "Singapore",
    publisher = "Association for Computational Linguistics",
    doi = "10.18653/v1/2023.findings-emnlp.243",
    pages = "3730--3748",
}

@inproceedings{kotek2023gender,
author = {Kotek, Hadas and Dockum, Rikker and Sun, David},
title = {Gender bias and stereotypes in Large Language Models},
year = {2023},
isbn = {9798400701139},
publisher = {Association for Computing Machinery},
address = {New York, NY, USA},
doi = {10.1145/3582269.3615599},
booktitle = {Proceedings of The ACM Collective Intelligence Conference},
pages = {12–24},
numpages = {13},
location = {Delft, Netherlands},
series = {CI '23}
}

@article{ecker2024misinformation,
	title = {Why misinformation must not be ignored},
	volume = {80},
	issn = {1935-990X},
	doi = {10.1037/amp0001448},
	number = {6},
	journal = {American Psychologist},
	author = {Ecker, Ullrich K. H. and Tay, Li Qian and Roozenbeek, Jon and van der Linden, Sander and Cook, John and Oreskes, Naomi and Lewandowsky, Stephan},
    year={2024},
	pages = {867--878},
}

@inproceedings{tang2024gendercare,
author = {Tang, Kunsheng and Zhou, Wenbo and Zhang, Jie and Liu, Aishan and Deng, Gelei and Li, Shuai and Qi, Peigui and Zhang, Weiming and Zhang, Tianwei and Yu, NengHai},
title = {GenderCARE: A Comprehensive Framework for Assessing and Reducing Gender Bias in Large Language Models},
year = {2024},
isbn = {9798400706363},
publisher = {Association for Computing Machinery},
address = {New York, NY, USA},
doi = {10.1145/3658644.3670284},
booktitle = {Proceedings of the 2024 on ACM SIGSAC Conference on Computer and Communications Security},
pages = {1196–1210},
numpages = {15},
location = {Salt Lake City, UT, USA},
series = {CCS '24}
}

@incollection{vanmassenhove2024gender,
  title={Gender bias in machine translation and the era of large language models},
  author={Vanmassenhove, Eva},
  booktitle={Gendered Technology in Translation and Interpreting},
  pages={225--252},
  year={2024},
  publisher={Routledge}
}

@article{bajaj2024evaluating,
  title={Evaluating gender bias of LLMs in making morality judgements},
  author={Bajaj, Divij and Lei, Yuanyuan and Tong, Jonathan and Huang, Ruihong},
  journal={arXiv preprint arXiv:2410.09992},
  year={2024}
}

@article{radaideh2025fairness,
title = {Fairness and social bias quantification in Large Language Models for sentiment analysis},
journal = {Knowledge-Based Systems},
volume = {319},
pages = {113569},
year = {2025},
issn = {0950-7051},
doi = {10.1016/j.knosys.2025.113569},
author = {Mohammed I. Radaideh and O. Hwang Kwon and Majdi I. Radaideh},
publisher={Elsevier}
}

@INPROCEEDINGS{huang2025unmasking,
  author={Huang, Tianyi and Yi, Jingyuan and Yu, Peiyang and Xu, Xiaochuan},
  booktitle={2025 8th International Conference on Advanced Algorithms and Control Engineering (ICAACE)}, 
  title={Unmasking Digital Falsehoods: A Comparative Analysis of LLM-Based Misinformation Detection Strategies}, 
  year={2025},
  volume={},
  number={},
  pages={2470-2476},
  doi={10.1109/ICAACE65325.2025.11020217}}

@ARTICLE{kuntur2024under,
  author={Kuntur, Soveatin and Wróblewska, Anna and Paprzycki, Marcin and Ganzha, Maria},
  journal={IEEE Transactions on Artificial Intelligence}, 
  title={Under the Influence: A Survey of Large Language Models in Fake News Detection}, 
  year={2025},
  volume={6},
  number={2},
  pages={458-476},
  doi={10.1109/TAI.2024.3471735}}

@inproceedings{sobhani2024towards,
    title = "Towards Fairer {NLP} Models: Handling Gender Bias In Classification Tasks",
    author = "Sobhani, Nasim  and
      Delany, Sarah",
    editor = "Fale{\'n}ska, Agnieszka  and
      Basta, Christine  and
      Costa-juss{\`a}, Marta  and
      Goldfarb-Tarrant, Seraphina  and
      Nozza, Debora",
    booktitle = "Proceedings of the 5th Workshop on Gender Bias in Natural Language Processing (GeBNLP)",
    month = aug,
    year = "2024",
    address = "Bangkok, Thailand",
    publisher = "Association for Computational Linguistics",
    doi = "10.18653/v1/2024.gebnlp-1.10",
    pages = "167--178",
}

@inproceedings{russo2025tracing,
author = {Russo, Mayra and Merenda, Flavio and Gomez-Perez, Jose Manuel and Vidal, Maria-Esther},
title = {Tracing Bias for Fairer Content-Based Misinformation Detection},
year = {2025},
isbn = {9798400713316},
publisher = {Association for Computing Machinery},
address = {New York, NY, USA},
doi = {10.1145/3701716.3717534},
booktitle = {Companion Proceedings of the ACM on Web Conference 2025},
pages = {2670–2679},
numpages = {10},
location = {Sydney NSW, Australia},
series = {WWW '25}
}

@inproceedings{dacon2021does,
author = {Dacon, Jamell and Liu, Haochen},
title = {Does Gender Matter in the News? Detecting and Examining Gender Bias in News Articles},
year = {2021},
isbn = {9781450383134},
publisher = {Association for Computing Machinery},
address = {New York, NY, USA},
doi = {10.1145/3442442.3452325},
booktitle = {Companion Proceedings of the Web Conference 2021},
pages = {385–392},
numpages = {8},
location = {Ljubljana, Slovenia},
series = {WWW '21}
}

@article{weerts2023fairlearn,
  author  = {Hilde Weerts and Miroslav DudÃ­k and Richard Edgar and Adrin Jalali and Roman Lutz and Michael Madaio},
  title   = {Fairlearn: Assessing and Improving Fairness of AI Systems},
  journal = {Journal of Machine Learning Research},
  year    = {2023},
  volume  = {24},
  number  = {257},
  pages   = {1--8},
}

@inproceedings{ouyang2022training,
 author = {Ouyang, Long and Wu, Jeffrey and Jiang, Xu and Almeida, Diogo and Wainwright, Carroll and Mishkin, Pamela and Zhang, Chong and Agarwal, Sandhini and Slama, Katarina and Ray, Alex and Schulman, John and Hilton, Jacob and Kelton, Fraser and Miller, Luke and Simens, Maddie and Askell, Amanda and Welinder, Peter and Christiano, Paul F and Leike, Jan and Lowe, Ryan},
 booktitle = {Advances in Neural Information Processing Systems},
 editor = {S. Koyejo and S. Mohamed and A. Agarwal and D. Belgrave and K. Cho and A. Oh},
 pages = {27730--27744},
 publisher = {Curran Associates, Inc.},
 title = {Training language models to follow instructions with human feedback},
 volume = {35},
 year = {2022}
}

@misc{ganguli2022red,
      title={Red Teaming Language Models to Reduce Harms: Methods, Scaling Behaviors, and Lessons Learned}, 
      author={Deep Ganguli and Liane Lovitt and Jackson Kernion and Amanda Askell and Yuntao Bai and Saurav Kadavath and Ben Mann and Ethan Perez and Nicholas Schiefer and Kamal Ndousse and Andy Jones and Sam Bowman and Anna Chen and Tom Conerly and Nova DasSarma and Dawn Drain and Nelson Elhage and Sheer El-Showk and Stanislav Fort and Zac Hatfield-Dodds and Tom Henighan and Danny Hernandez and Tristan Hume and Josh Jacobson and Scott Johnston and Shauna Kravec and Catherine Olsson and Sam Ringer and Eli Tran-Johnson and Dario Amodei and Tom Brown and Nicholas Joseph and Sam McCandlish and Chris Olah and Jared Kaplan and Jack Clark},
      year={2022},
      eprint={2209.07858},
      archivePrefix={arXiv},
      primaryClass={cs.CL},
}

@article{chalehchaleh2025addressing,
  title={Addressing data scarcity in multilingual fake news detection: an LLM-based dataset augmentation approach},
  author={Chalehchaleh, Razieh and Farahbakhsh, Reza and Crespi, Noel},
  journal={Social Network Analysis and Mining},
  volume={15},
  number={1},
  pages={1--16},
  year={2025},
  doi     = {10.1007/s13278-025-01505-z},
  publisher={Springer}
}
\end{document}